\documentclass[conference]{IEEEtran}
\IEEEoverridecommandlockouts
\usepackage{cite}
\usepackage{amsmath,amssymb,amsfonts}
\usepackage{algorithm,algorithmic}
\usepackage{graphicx}
\usepackage{textcomp}
\usepackage{xcolor}
\def\BibTeX{{\rm B\kern-.05em{\sc i\kern-.025em b}\kern-.08em
    T\kern-.1667em\lower.7ex\hbox{E}\kern-.125emX}}
    
\usepackage{fancyhdr}
\usepackage{balance}
\usepackage{subfigure}
\usepackage{bm}
\usepackage{multirow}
\usepackage{booktabs}
\usepackage[bookmarks=false]{hyperref}
\usepackage{pdfrender}
\newcommand*{\boldcheckmark}{
  \textpdfrender{
    TextRenderingMode=FillStroke,
    LineWidth=.5pt,
  }{\checkmark}
}
\newcommand{\R}{\mathbb{R}}

\begin{document}

\title{SpineOne: A One-Stage Detection Framework for Degenerative Discs and Vertebrae \\

\thanks{This research was done when Jiabo He interned with the DAMO Academy, Alibaba Group. A short version of this paper has been accepted by IEEE BIBM 2021. Correspondence to Xingjun Ma and Yu Wang.}
}

\author{
\IEEEauthorblockN{Jiabo He\textsuperscript{1,2}, Wei Liu\textsuperscript{2}, Yu Wang\textsuperscript{2}, Xingjun Ma\textsuperscript{3}, Xian-Sheng Hua\textsuperscript{2}}
\IEEEauthorblockA{\textsuperscript{1}\textit{School of Computing and Information Systems, The University of Melbourne, Australia} \\
\textsuperscript{2}\textit{DAMO Academy, Alibaba Group, China} \\
\textsuperscript{3}\textit{School of Information Technology, Deakin University, Australia} \\
jiaboh@student.unimelb.edu.au, daniel.ma@deakin.edu.au, \{vivi.lw, tonggou.wangyu, xiansheng.hxs\}@alibaba-inc.com}
}

\maketitle

\begin{abstract}
Spinal degeneration plagues many elders, office workers, and even the younger generations. Effective pharmic or surgical interventions can help relieve degenerative spine conditions. However, the traditional diagnosis procedure is often too laborious. Clinical experts need to detect discs and vertebrae from spinal magnetic resonance imaging (MRI) or computed tomography (CT) images as a preliminary step to perform pathological diagnosis or preoperative evaluation. Machine learning systems have been developed to aid this procedure generally following a two-stage methodology: first perform anatomical localization, then pathological classification. Towards more efficient
and accurate diagnosis, we propose a one-stage detection framework termed SpineOne to simultaneously localize and classify degenerative discs and vertebrae from MRI slices. SpineOne is built upon the following three key techniques: 1) a new design of the keypoint heatmap to facilitate simultaneous keypoint localization and classification; 2) the use of attention modules to better differentiate the representations between discs and vertebrae; and 3) a novel gradient-guided objective association mechanism to associate multiple learning objectives at the later training stage. Empirical results on the Spinal Disease Intelligent Diagnosis Tianchi Competition (SDID-TC) dataset of 550 exams demonstrate that our approach surpasses existing methods by a large margin.
\end{abstract}


\section{Introduction}
\label{sec:introduction}
\begin{figure}[t]
\centering
\includegraphics[width=1\columnwidth]{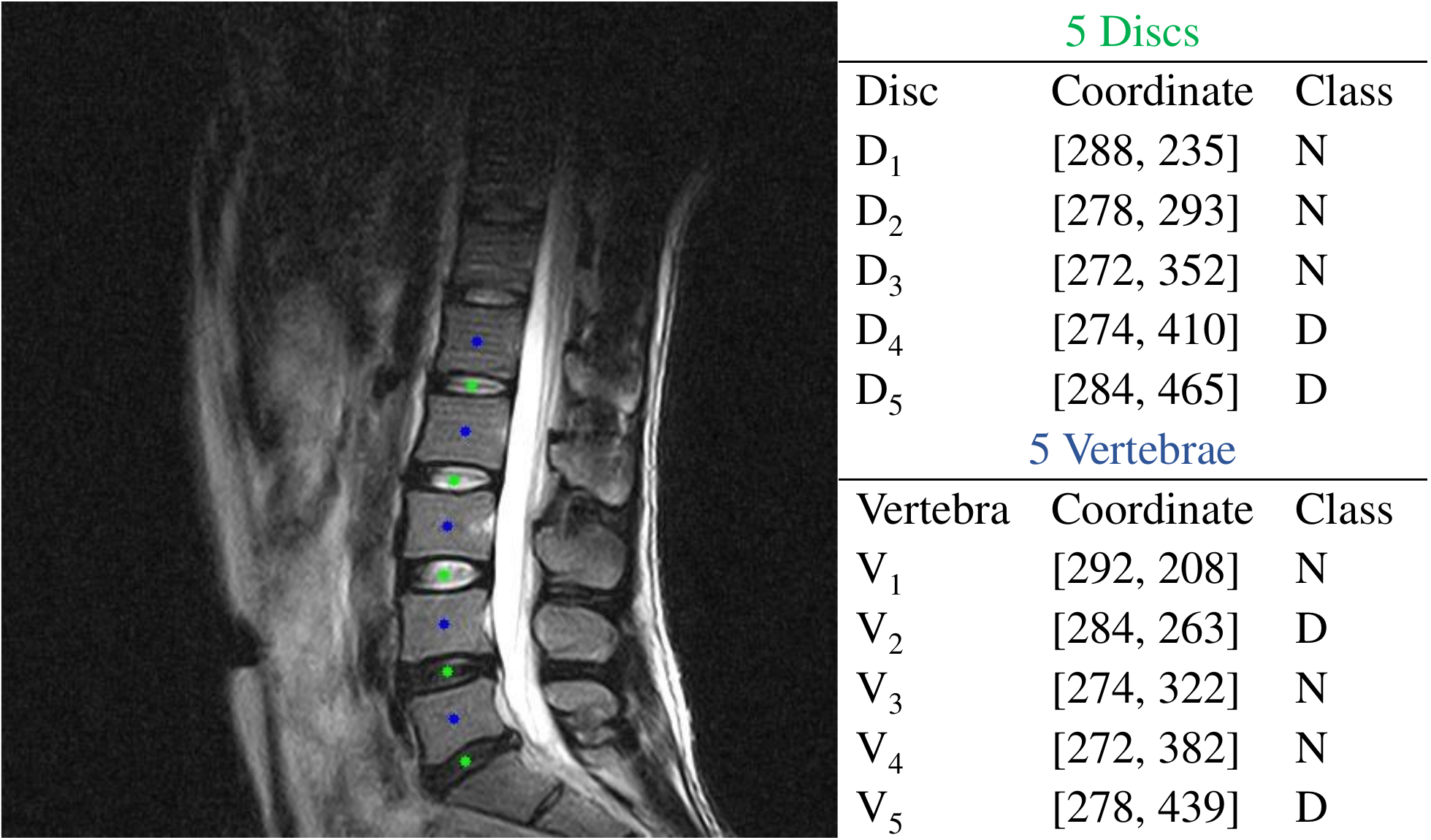}
\caption{Illustration of 5 discs (centroids in green) and 5 vertebrae (centroids in blue) from the MRI slice of the lumbar spine. $D_1$, $D_2$, $D_3$, $D_4$, and $D_5$ denote the disc $L_1$-$L_2$, $L_2$-$L_3$, $L_3$-$L_4$, $L_4$-$L_5$, and $L_5$-$S_1$, respectively; $V_1$, $V_2$, $V_3$, $V_4$, and $V_5$ denote the vertebra $L_1$, $L_2$, $L_3$, $L_4$, and $L_5$, respectively; where $L_{1-5}$ are the five lumbar levels from top to bottom and $S_1$ is the sacral spine. N and D denote Normal and Degenerative classes.}
\label{spine}
\end{figure}

\IEEEPARstart{S}{pinal} diseases have become increasingly common nowadays, among which the degeneration of spines is nearly inevitable with aging. Degenerative spine conditions involve the gradual loss of normal structures and functions over time, which may be caused by aging, tumors, infections and arthritis \cite{ucdavis}. Magnetic resonance imaging (MRI) and computed tomography (CT) techniques are used to visualize the anatomical structures of the spine before pathological diagnosis and treatment. The lumbar spine consists of $5$ discs and $5$ vertebrae (Fig. \ref{spine}). Discs act as shock absorbers between vertebrae and both of them can involve degenerative conditions \cite{NY}.

Degenerative spine conditions can be relieved by pharmic or surgical interventions. Such labor-intensive interventions rely on clinical experts to manually identify degenerative discs and vertebrae from spinal MRI or CT images. Traditional image processing systems have been developed to help localize discs and vertebrae via an automated process \cite{kadoury2013spine, chen2015localization, korez2015framework}. Convolutional neural networks (CNNs) have also been introduced to detect and segment discs and vertebrae from spinal MRIs \cite{jamaludin2016spinenet, han2018spine, levine2019automatic}.
For example, SpineNet generates slice-level (i.e., image-level) radiological scores for discs \cite{jamaludin2016spinenet} while Spine-GAN segments discs, vertebrae, and neural foramen with pixel-level results \cite{han2018spine}. Different from above existing works, we localize discs and vertebrae by predicting the locations of their centroids (i.e., keypoints), as well as generating their anatomical-structure-level (i.e., keypoint-level) degenerative categories. We will further show that our task enables minimal annotation effort from experts compared with existing segmentation tasks.

There are three main challenges of applying existing methods to our task: 1) there are multiple centroids ($5$ discs and $5$ vertebrae) and multiple pathological classes ($2$ classes for either discs or vertebrae); 2) discs and vertebrae need to be diagnosed simultaneously, despite their high inter-structural similarity and intra-structural variation; and 3) there are multiple learning objectives including localization and classification (for each disc and vertebra) and the localization involves both keypoint localization and offset regression.
To address these challenges, we propose \textit{SpineOne}, an efficient one-stage framework that can automatically localize and classify degenerative discs and vertebrae from MRIs. SpineOne is more friendly to clinical experts than two-stage frameworks, since it can provide the localization and classification results at the same time. SpineOne takes MRI slices of a lumbar spine as inputs and outputs centroids and offsets for detected discs and vertebrae, and at the same time, outputs the pixel-wise probability of each disc or vertebra being degenerative. Therefore, SpineOne is able to help clinical experts relieve from large amounts of the workload
by serving as an assistant, making efficient and accurate
diagnosis.

Specifically, based on CNNs, SpineOne addresses the above three challenges using three novel techniques. \textbf{First}, a one-channel-per-class (OCPC) keypoint heatmap is designed to promote simultaneous localization and classification without introducing additional heads or output channels into the network. Our OCPC design is advantageous as the centroids of discs and vertebrae are spatially separated following physiological rigidity, which can help capture the geometrical and classification correlations among keypoints (Fig. \ref{spine}).
\textbf{Second}, we introduce dual self-attention modules to adaptively integrate similar features at different scales -- an inspiration from human experts who would inspect MRI slices from a global view with visual acuity. Specifically, we use 1) a position attention module (PAM) to aggregate the feature at each position by a weighted sum of features over all positions of the image, and 2) a channel attention module (CAM) to integrate features across channels. 
\textbf{Third}, we introduce a novel gradient-guided objective association (OA) mechanism to adaptively associate two types of objectives, i.e., using the gradient of the loss w.r.t. the heatmap to guide the learning of the offset. OA is a generic technique to boost the learning of the main objectives in multi-head models at the later training stage. 

In summary, our main contributions are:
\begin{itemize}
\item We propose a novel one-channel-per-class (OCPC) keypoint heatmap design to facilitate simultaneous keypoint localization and classification. Our OCPC heatmap compacts all keypoints of the same class into the same channel given they are spatially separated.
\item We introduce dual self-attention modules for the learning of more distinguishable representations between discs and vertebrae: a position attention module (PAM) to model inner-image spatial interaction, and a channel attention module (CAM) to model inter-channel correlation.
\item We propose a novel gradient-guided objective association (OA) mechanism to explicitly connect heatmap learning with offset learning. This makes the learning of the heatmap more effective, which is the main objective of keypoint localization and classification.
\item We integrate above three techniques into one novel one-stage detection framework SpineOne, and empirically show, on the Spinal Disease Intelligent Diagnosis Tianchi Competition dataset \cite{tianchi}, that SpineOne can outperform the current state-of-the-art methods by a large margin.
\end{itemize}

\section{Related Work}
In this section, we first review existing deep learning models developed for spine-related tasks. We then discuss various state-of-the-art methods for keypoint detection.

\subsection{Segmentation, Detection and Pathological Classification on Spines}
A number of deep learning models have been proposed for spine-related tasks including anatomical segmentation \cite{sekuboyina2017attention, zheng2017evaluation, pang2020spineparsenet}, detection \cite{levine2019automatic, glocker2012automatic, chen2015automatic, zhao2021automatic}, and pathological classification \cite{chmelik2018deep, jamaludin2016spinenet} from either MRIs or CT images. These works have provided great assistance to clinical experts in osteoporosis assessment, fractures detection and aging process analysis. There have also been several multi-task learning studies on spines. For example, two-stage methods have been proposed to train one network for vertebral segmentation and disc image extraction, and the other network for stenosis grading \cite{lu2018deepspine}.
Different from these two-stage methods, one-stage methods are able to segment discs, vertebrae, and neural foramen simultaneously using GAN-based models \cite{han2018spine}. The sequential conditional reinforcement learning network (SCRL) can also tackle the simultaneous vertebral detection and segmentation from MRIs \cite{zhang2021sequential}. 2D CNNs are frequently used to process MRI slices while 3D CNNs are popular for 3D CT images.

While there exists a body of work on spine-related tasks, our work is notably different. Specifically, our task requires to localize $5$ discs and $5$ vertebrae on the lumbar spine at the same time by localizing their centroids, which minimizes the annotation effort of experts. Furthermore, we classify the degenerative conditions for each disc and vertebra. To the best of our knowledge, none of the existing works has addressed the same task before.
In this work, we address this gap by proposing an efficient and accurate one-stage framework.
\begin{figure*}[pt]
\centering
\includegraphics[width=1.9\columnwidth]{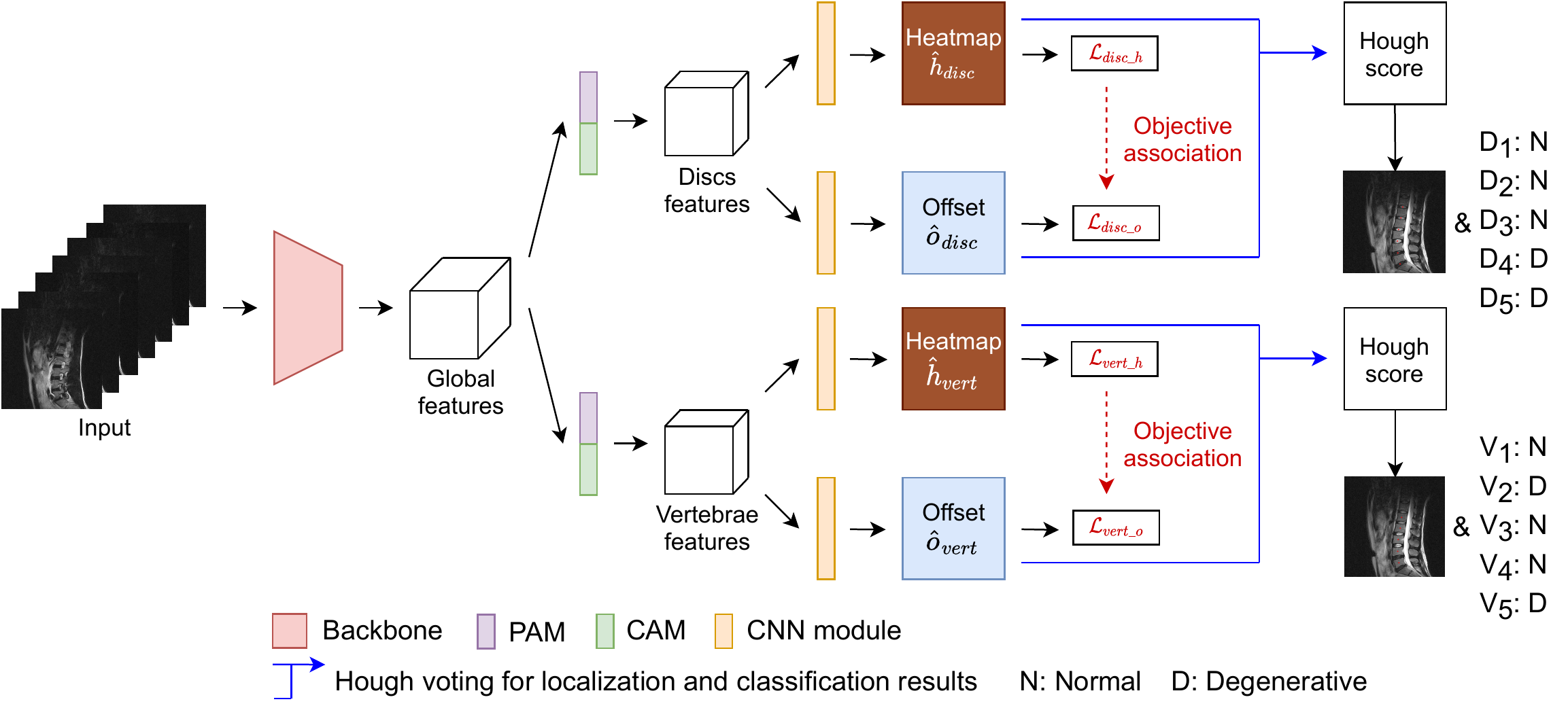}
\caption{Overview of our proposed framework SpineOne. One example output of our framework can be found in Fig. \ref{spine}. The structures of PAM and CAM modules are shown in Fig. \ref{image:attention_module}.}
\label{framework}
\end{figure*}

\subsection{Keypoint Detection}
Keypoint detection is fundamental to numerous vision tasks \cite{rashid2017interspecies, simon2017hand, georgakis2018end}, amongst which human pose estimation and keypoint-based object detection are the two most related topics to our task.

\subsubsection{Human pose estimation}
Single-person pose estimation localizes human anatomical keypoints/parts by regressing either spatial joint coordinates \cite{toshev2014deeppose} or location heatmaps \cite{tompson2014joint, wei2016convolutional}. Multi-person pose estimation can be achieved via either bottom-up \cite{cao2017realtime, hidalgo2019single} or top-down \cite{fang2017rmpe, chen2018cascaded} approaches with fully CNNs. Besides fully CNNs, specialized CNN architectures were also proposed for keypoint detection. For example, the stacked hourglass network repeats the bottom-up and top-down processing in conjunction with intermediate supervision \cite{newell2016stacked}. The HRNet maintains high-resolution representations through the entire training process using high-to-low resolution sub-networks \cite{sun2019deep}. Keypoint detection requires rich spatial correlations and contextual information captured from both high- and low-level representations \cite{yang2017learning, chu2017multi}. 
As such, keypoint detection can potentially boost other tasks via single-shot approaches, e.g., instance segmentation, semantic segmentation, and even image parsing (i.e., panoptic segmentation) \cite{papandreou2018personlab, chen2018encoder, yang2019deeperlab}.
In this work, state-of-the-art architectures for keypoint
detection are leveraged at the first stage of two-stage baselines,
followed by classifiers for those positive/detected keypoints at the second stage.

\subsubsection{Keypoint-based object detection}
\begin{figure}[t]
    \centering
    \subfigure[Position attention module (PAM)]
    {\includegraphics[width=1\columnwidth]{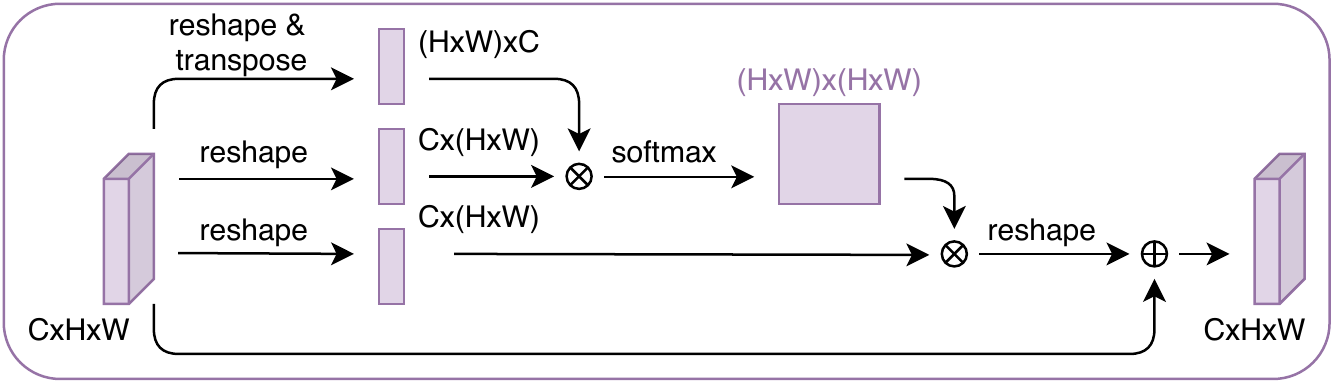}}
    \subfigure[Channel attention module (CAM)]
    {\includegraphics[width=1\columnwidth]{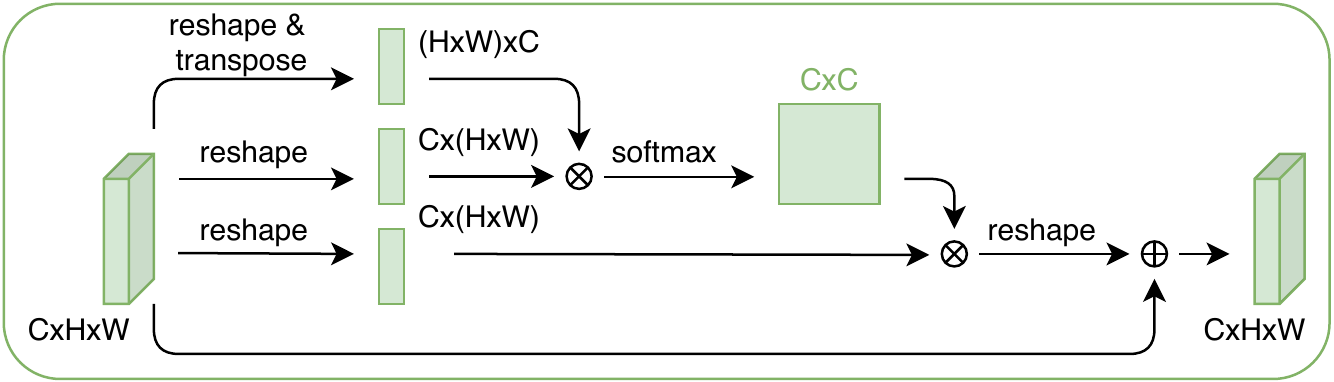}}
    \caption{Structure details of PAM and CAM. $\otimes$: matrix multiplication; $\oplus$: element-wise addition.}
    \label{image:attention_module}
\end{figure}

These methods detect objects based on the keypoint heatmaps by one-stage, anchor-free detectors. CornetNet \cite{law2018cornernet}, CenterNet1 \cite{duan2019centernet}, ExtremeNet \cite{zhou2019bottom}, and CentripetalNet \cite{dong2020centripetalnet} predict locations of keypoints (corners, centroids, or extreme points) and group them into same bounding boxes if they are geometrically aligned. Compared with these works, we only need to localize the centroids of discs and vertebrae and no grouping is needed. Several works address the simultaneous localization and classification problem by generating pixel-wise results. CenterNet2 estimates pixel-level categories of objects along with their sizes and offsets \cite{zhou2019objects}. FCOS generates pixel-wise classification, center-ness, and bounding box (top, down, left, right) results using multi-head CNNs \cite{tian2019fcos}. In our task, it is also required to simultaneously localize and classify degenerative conditions of each disc and vertebra. We tackle this challenge by introducing a novel keypoint heatmap to these architectures.

\section{Proposed One-stage Approach}
\label{section:approach}
\noindent\textbf{Problem definition.}
Let $\bm{I} \in \mathbb{R}^{n \times H \times W}$ be the sagittal T2 sequence of an input exam with $n$ slices, height $H$, and width $W$. SpineOne outputs the keypoint heatmaps ($\bm{\hat{h}}_{disc}$ and $\bm{\hat{h}}_{vert}$) and the corresponding offset maps ($\bm{\hat{o}}_{disc}$ and $\bm{\hat{o}}_{vert}$) for both discs and vertebrae, respectively. The goal is to design and train SpineOne to output consistent $\bm{\hat{h}}_{disc}$, $\bm{\hat{h}}_{vert}$, $\bm{\hat{o}}_{disc}$, and $\bm{\hat{o}}_{vert}$ with the ground truth $\bm{h}_{disc}$, $\bm{h}_{vert}$, $\bm{o}_{disc}$, and $\bm{o}_{vert}$. The ground truth can be generated from the annotations (locations and classes). After postprocessing above outputs, we compute the \textit{precision}, \textit{recall}, and \textit{F1 score} to evaluate the final performance of SpineOne comprehensively.

\subsection{Framework Overview}
An ordinary detector has at least three components, i.e., input, backbone, and head \cite{bochkovskiy2020yolov4}. Recent detectors also insert a neck component between the backbone and the head. For clarity, we take the backbone along with the neck (if there is one) as a whole module for feature extraction, fusion, and learning. The proposed SpineOne framework is illustrated in Fig. \ref{framework}. 
For either the discs (top) or vertebrae (bottom) branch, traditional two-stage frameworks use sequential output heads with one for localization and the other for classification. SpineOne simplifies this process by using one single head (the 'Heatmap' output in Fig. \ref{framework}) for two purposes, i.e., keypoint localization and classification. Note that it still needs one additional head for the offset learning, respectively.

We select $n=7$ slices in the middle of the sagittal T2 sequence of MRIs as inputs, which are the main focus of clinical experts when diagnosing degenerative spines \cite{jamaludin2016spinenet}. A backbone is used to extract the global feature map $\bm{U}_{global}$, which is then forked by two dual self-attention modules (i.e., PAM and CAM) to learn discs and vertebrae feature representations ($\bm{U}_{disc}$ and $\bm{U}_{vert}$). The two feature representations are then separately fed into CNN modules to generate keypoint heatmaps ($\bm{\hat{h}}_{disc}$, and $\bm{\hat{h}}_{vert}$), and corresponding offset maps ($\bm{\hat{o}}_{disc}$ and $\bm{\hat{o}}_{vert}$), which is done for discs and vertebrae in parallel. Details of the framework structure is accessible to readers in Table \ref{table:framework}.
Loss functions are defined between outputs (i.e., $\bm{\hat{h}}_{disc}$, $\bm{\hat{h}}_{vert}$, $\bm{\hat{o}}_{disc}$, and $\bm{\hat{o}}_{vert}$) and their ground truth (i.e., $\bm{h}_{disc}$, $\bm{h}_{vert}$, $\bm{o}_{disc}$, and $\bm{o}_{vert}$) \cite{he2020learning, he2021alpha}.
Note that we leverage the gradient of the heatmap loss w.r.t. the heatmap to guide the learning of the offset by the objective association (OA) mechanism. The entire network is trained in an end-to-end and supervised fashion. There is also a postprocessing step to produce localization and classification results with Hough voting.
\begingroup
\begin{figure}[tp]
\setlength{\tabcolsep}{2pt}
\centering
\begin{tabular}{cccc}
\includegraphics[width=0.22\columnwidth]{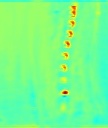}&
\includegraphics[width=0.22\columnwidth]{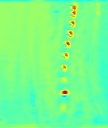}&
\includegraphics[width=0.22\columnwidth]{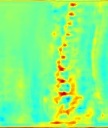}&
\includegraphics[width=0.22\columnwidth]{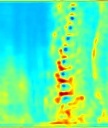}
\\
\includegraphics[width=0.22\columnwidth]{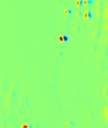}&
\includegraphics[width=0.22\columnwidth]{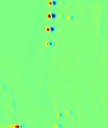}&
\includegraphics[width=0.22\columnwidth]{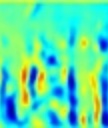}&
\includegraphics[width=0.22\columnwidth]{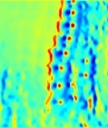}
\\
(a) Discs & (b) Discs & (c) Vertebrae & (d) Vertebrae
\end{tabular}
    \caption{Visualization of feature maps learned by our framework w/ (top) or w/o (bottom) attention modules. Those features are randomly selected channels of discs features and vertebrae features.}
    \label{image:attention_visual}
\end{figure}
\endgroup

\subsection{Attention Modules}
We introduce attention modules for capturing the long-range dependencies among input slices \cite{vaswani2017attention, wang2018non, woo2018cbam, liu2019end, zhang2019self}. Specifically, we use the dual self-attention modules \cite{fu2019dual} to learn more discriminative disc and vertebra representations. The dual self-attention modules consist of a position attention module (PAM) and a channel attention module (CAM). PAM selectively aggregates the feature at each position by a weighted sum of features at all positions within the image (Fig. \ref{image:attention_module}a), while CAM can notice interdependent channel maps by integrating associated features among all channels (Fig. \ref{image:attention_module}b).
Note that the attention maps of PAM and CAM are different in size as feature matrixes are multiplied in reversed orders. It is worth mentioning that PAM is memory hungry as there can be $\sim$billions of elements in the intermediate attention map of size $(H\times W)\times(H\times W)$ (e.g., there are approximately $0.3$ billion elements when $H=W=128$). Fig. \ref{image:attention_visual} provides a visual comparison between feature maps learned w/ and w/o dual attention modules from MRIs. With attention modules (the top row), the network is able to accurately locate the discs and vertebrae, and pays more attention on the lumbar spine. Moreover, the attention module for discs mainly focuses on regions within discs whereas the module for vertebrae focuses on both regions within vertebrae and their surrounding regions. 
However, without attention modules, the network may fail to locate discs nor vertebrae, and the attention may be shifted towards unimportant regions. We will also show in our experiments that the final performance can drop considerably without attention modules.

\subsection{OCPC Keypoint Heatmap}
\begin{figure}[tp]
\centering
\includegraphics[width=1\columnwidth]{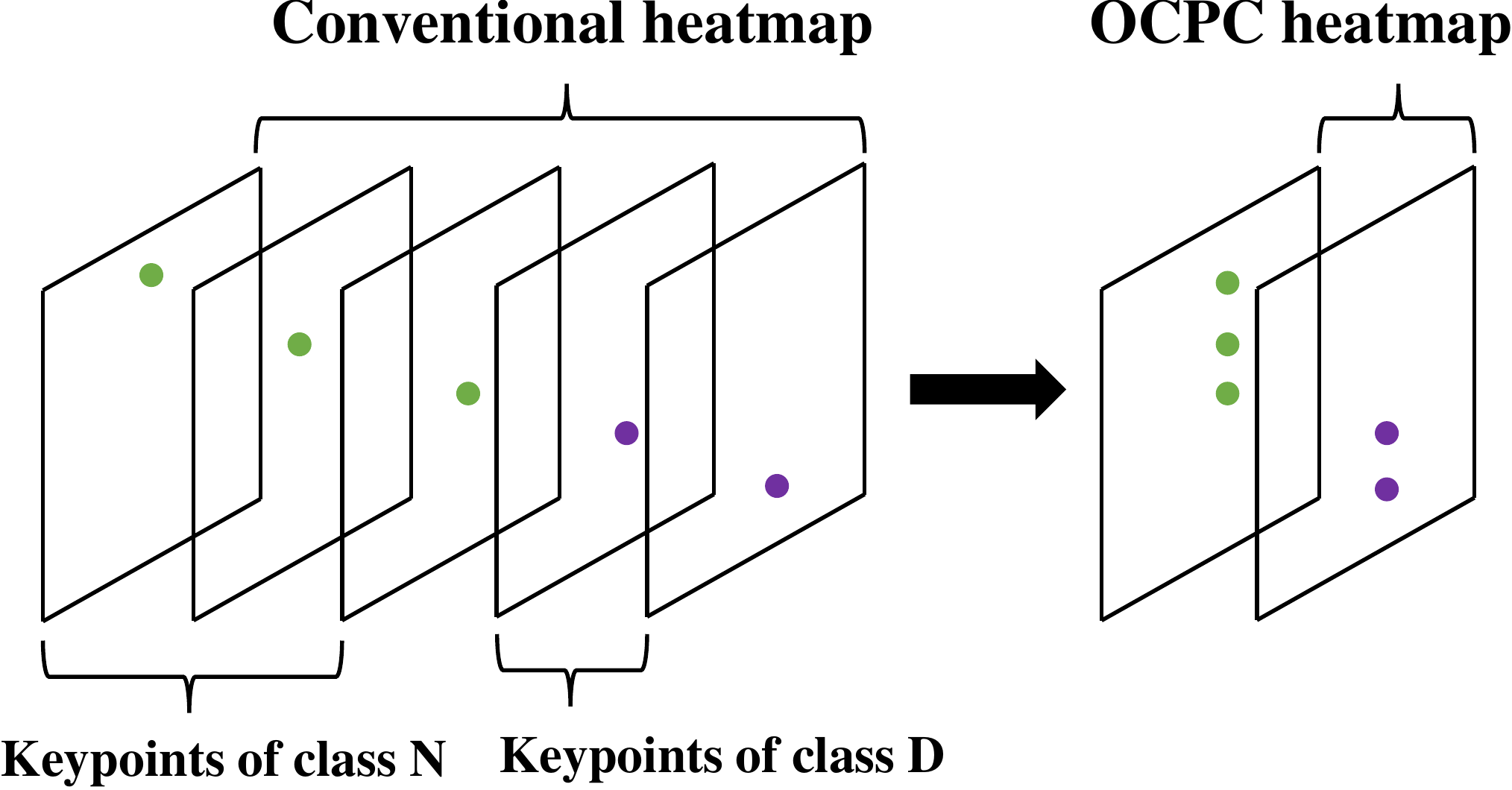}
\caption{Comparison between the conventional heatmap and our proposed one-channel-per-class (OCPC) heatmap. Different colors of keypoints represent different classes.}
\label{image:ocpc}
\end{figure}

The conventional keypoint heatmap has only one positive (i.e., 1) value indicating the location of the ground truth keypoint, while all other locations are zeros. For example, if the heatmap is in size of $640\times 640$, there will be $409,599$ zeros and only $1$ value of one. The foreground/background ratio is then $1/409599$ for one keypoint. In order to alleviate the dominance of the background, conventional approaches generate either Gaussian heatmaps \cite{he2017mask, newell2017associative} or binary heatmaps \cite{papandreou2017towards, papandreou2018personlab}, which involve the processing of local regions (within a radius $R$) around keypoints and the extension to multiple channels with one channel for each keypoint of each class. This basically means that, for an object with $K$ keypoints and $C$ classes, a heatmap with $K+C$ channels will be generated.
These methods force the network to have either redundant output heads or heads with redundant channels, resulting in a complex network architecture that is hard to train. As such, these designs are not suitable for medical applications where a general lack of data hinders the training of complex models. Alternatively, we propose to convert the conventional heatmap to a one-channel-per-class (OCPC) keypoint heatmap (Fig. \ref{image:ocpc}), which not only compacts channels of heads within $C=2$ channels (the normal class versus the degenerative class), but also captures the geometrical and classification correlations among keypoints.

Let $\bm{x}$ be the middle slice of the sagittal T2 sequence, with $x_i$ ($i=1, \dots, N$) being the positional index of the slice. $N$ is the total number of pixels. Let $y_k$ be the $k^{th}$ keypoint, and $\mathcal{D}_R(y_k)=\{x_i:||x_i-y_k|| \leq R \}$ be a disk of radius $R$ centered of $y_k$. 
For each keypoint $y_k$ of class $c$, its OCPC heatmap is defined as follows:
\begin{equation}
    p_{k,c}(x_i) =
    \begin{cases}
     1, \;\; \text{if} \; x_i \in \mathcal{D}_R(y_k), \\
     0, \;\; \text{otherwise.}
    \end{cases}
\end{equation}
This will produce a dense binary OCPC heatmap with $C=2$ channels for all keypoints with integrated class information.
We set $R$ to 6 pixels (i.e., $0.4375 \times 6=2.625mm$ after spacing alignment and resizing all slices to $640\times 640$ pixels) such that disks do not overlap with each other.
The foreground/background ratio thus increases from $1/409599$ to approximately $1/4403$ for one keypoint, which can further increase with multiple keypoints of the same class generated in the same channel.
OCPC is a compact and generic heatmap design that can be readily applied to any applications where the keypoints are spatially separated from each other.

\subsection{Offset and Hough Voting}
\subsubsection{Short-range offset}
There are three types of offsets to be considered: 1) short-range, 2) mid-range, and 3) long-range offsets \cite{papandreou2018personlab}.
In our framework, we adopt the short-range offset to improve the keypoint localization. 
The mid-range offset associates keypoints of the same instance while the long-range offset calculates distances between the keypoint and all points within its instance for other tasks, e.g., instance segmentation. We thus do not consider the latter two offsets in our task.
The short-range offset with $2C=4$ channels ($2$ channels of coordinate $x$ and $y$ for each keypoint of each class) that points from an image position $x_i$ to the $k^{th}$ keypoint of class $c$ is defined as:
\begin{equation}
S_{k,c}(x_i)=y_{k}-x_i, \; x_i \in \mathcal{D}_R(y_k).
\label{offset}
\end{equation}

\subsubsection{Hough voting}
We further aggregate the keypoint heatmap and the short-range offset map into a Hough score map $h_{k,c}(x)$ for each class $c$ \cite{papandreou2018personlab}:
\begin{equation}
h_{k,c}(x)=\frac{1}{\pi R^2}\sum_{i=1}^{N}p_{k,c}(x_i)B(x_i+S_{k,c}(x_i)-x),
\label{hough}
\end{equation}
where $p_{k,c}(x_i)$ is the sigmoid probability in channel $c$ in the heatmap and $B(\cdot)$ is the bilinear interpolation kernel for size match between outputs and the ground truth.
We sort and select the top five points with the top five highest scores in the Hough score map as the prediction results for discs and vertebrae, respectively.
The Hough score map facilitates accurate keypoint localization by considering both keypoint locations and their surrounding pixels. We refer readers to \cite{papandreou2018personlab} for more details of the Hough voting.

\subsection{Gradient-Guided Objective Association}
\label{section:gradient}
Our SpineOne framework has $4$ output heads with two heads for discs and the other two for vertebrae. We combine the loss functions defined for each head into one total loss:
\begin{equation}
\begin{aligned}
\mathcal{L} = \mathcal{L}_{disc\_h}+\mathcal{L}_{disc\_o}+\mathcal{L}_{vert\_h}+\mathcal{L}_{vert\_o},
\end{aligned}
\label{loss}
\end{equation}
where the first two terms are for discs (i.e., $\mathcal{L}_{disc\_h}$ for keypoint heatmap and $\mathcal{L}_{disc\_o}$ for offset) and the last two terms are for vertebrae (i.e., $\mathcal{L}_{vert\_h}$ and $\mathcal{L}_{vert\_o}$). 
We test different sets of hyper-parameters and find no noticeable improvement based on the above loss. The four loss terms are thus naturally balanced in our task and introducing hyper-parameters into the above loss is not necessary.
Regarding the exact forms of the loss functions, we use the focal loss ($FL(\hat{p})=-(1-\hat{p})^{\gamma}\textrm{log}(\hat{p}), \; \gamma>0$) \cite{lin2017focal} for heatmap losses, and the $L_1$ loss ($L_1(\hat{r})=|\hat{r}-r|$) for offset losses.

One remaining problem is that the heatmap generation is in fact more important than the offset regression because the heatmap provides both localization and classification results while the offset is useful only to the localization task. As such, the learning of the heatmap should lead the learning of the offset.
To address this problem, we propose a gradient-guided \textit{objective association} (OA) mechanism that associates heatmap loss terms with offset loss terms. Taking discs for an example, its two loss terms are:
\begin{equation}
\begin{aligned}
\mathcal{L}_{disc}
    & = \mathcal{L}_{disc\_h}+\mathcal{L}_{disc\_o} \\
    & = \mathcal{L}_{disc\_h}(\bm{\hat{h}}_{disc}, \bm{h}_{disc}) + \mathcal{L}_{disc\_o}(\bm{\hat{o}}_{disc}, \bm{o}_{disc}),
\end{aligned}
\label{loss2}
\end{equation}
where $\bm{\hat{h}}_{disc}$ and $\bm{h}_{disc} \in \R^{C\times H\times W}$ are predicted and ground truth keypoint heatmaps respectively, while $\bm{\hat{o}}_{disc}$ and $\bm{o}_{disc} \in \R^{2C\times H\times W}$ are predicted and ground truth short-range offset maps respectively. $C$ is the total number of classes. 
The proposed OA mechanism associates the above two disc loss terms as follows:
\begin{equation}
\small
\begin{aligned}
\mathcal{L}_{disc} = 
    & \, \mathcal{L}_{disc\_h}(\bm{\hat{h}}_{disc}, \bm{h}_{disc}) + \\ 
    & \, \mathcal{L}_{disc\_o}(\bm{R}(\bm{1}\oplus\frac{\partial \mathcal{L}_{disc\_h}}{\partial \bm{\hat{h}}_{disc}}) \odot \bm{\hat{o}}_{disc}, \bm{o}_{disc}),
\end{aligned}
\label{loss3}
\end{equation}
where the constant tensor $\bm{1} \in \R^{C\times H\times W}$ and the gradient of the heatmap loss w.r.t. the heatmap ($\partial \mathcal{L}_{disc\_h}/\partial \bm{\hat{h}}_{disc}$) are element-wise added up as a type of weight to differentiate the varying importance of pixels in the heatmap.
$\bm{R}(\cdot)$ is a kernel that reshapes the size of the input from $\R^{C\times H\times W}$ to $\R^{2C\times H\times W}$ by duplicating
channels. $\odot$ denotes element-wise product. We clip each gradient value in $\partial \mathcal{L}_{disc\_h}/\partial \bm{\hat{h}}_{disc}$ to $[-100, 100]$, then normalize it to $[0, 1]$ for each channel.
Similarly, we also associate the two loss terms for vertebrae as follows:
\begin{equation}
\small
\begin{aligned}
\mathcal{L}_{vert} = 
    & \, \mathcal{L}_{vert\_h}(\bm{\hat{h}}_{vert}, \bm{h}_{vert}) + \\ 
    & \, \mathcal{L}_{vert\_o}(\bm{R}(\bm{1}\oplus\frac{\partial \mathcal{L}_{vert\_h}}{\partial \bm{\hat{h}}_{vert}}) \odot \bm{\hat{o}}_{vert}, \bm{o}_{vert}).
\end{aligned}
\label{loss4}
\end{equation}

\begin{table*}[pt]
\small
\caption{Experimental results of our framework compared with state-of-the-art frameworks on the SDID-TC dataset. All evaluation results are the macro mean of normal and degenerative classes. The top two best results in every column are \textbf{boldfaced}.}
\begin{center}
\begin{tabular}{c|c|ccc|ccc}
\toprule
\textbf{Framework} & \textbf{Backbone} & \multicolumn{3}{c|}{\textbf{Discs}} & \multicolumn{3}{c}{\textbf{Vertebrae}} \\
& & Recall & Precision & F1 & Recall & Precision & F1 \\
\midrule
\midrule
\multirow{4}{*}{Two-stage} & Hourglass-52+Res18 & 0.800 & 0.831 & 0.815 & 0.773 & 0.797 & 0.785 \\
& Hourglass-104+Res18 & 0.808 & 0.837 & 0.822 & 0.782 & 0.801 & 0.791 \\
& HRNet-W32+Res18 & 0.809 & 0.839 & 0.823 & 0.771 & 0.790 & 0.780 \\
& HRNet-W48+Res18 & 0.816 & 0.844 & 0.829 & 0.788 & 0.807 & 0.797 \\
\midrule
\midrule
\multirow{6}{*}{One-stage} & DenseNet121 w/ FPN & 0.802 & 0.836 & 0.818 & 0.755 & 0.778 & 0.766 \\ 
& DenseNet169 w/ FPN & 0.811 & 0.845 & 0.827 & 0.767 & 0.795 & 0.781 \\ 
& FCOS(Res101) & 0.814 & 0.848 & 0.830 & 0.770 & 0.798 & 0.783 \\ 
& DeepLabv3+(Res50) & 0.813 & 0.848 & 0.830 & 0.763 & 0.784 & 0.774 \\
& DeepLabv3+(Res101) & 0.819 & 0.855 & 0.836 & 0.774 & 0.795 & 0.785 \\
& CenterNet(Res101) & 0.817 & 0.851 & 0.833 & 0.774 & 0.802 & 0.788 \\
\midrule
\multirow{6}{*}{\textbf{SpineOne}} & DenseNet121 w/ FPN & 0.837 & 0.869 & 0.852 & 0.806 & 0.827 & 0.816 \\ 
& DenseNet169 w/ FPN & 0.843 & 0.875 & 0.858 & 0.816 & 0.837 & 0.827 \\ 
& FCOS(Res101) & 0.849 & 0.878 & 0.863 & 0.822 & 0.842 & 0.832 \\ 
& DeepLabv3+(Res50) & 0.847 & 0.873 & 0.860 & 0.811 & 0.834 & 0.823 \\
& DeepLabv3+(Res101) & \textbf{0.859} & \textbf{0.891} & \textbf{0.874} & \textbf{0.840} & \textbf{0.859} & \textbf{0.849} \\
& CenterNet(Res101) & \textbf{0.857} & \textbf{0.888} & \textbf{0.872} & \textbf{0.839} & \textbf{0.860} & \textbf{0.849} \\
\bottomrule
\end{tabular}
\label{table:result0}
\end{center}
\end{table*}

The gradient of the heatmap loss w.r.t. the heatmap highlights the importance of each pixel in the keypoint heatmap to both localization and classification tasks. Therefore, the OA mechanism encourages the network to focus more on important regions learned from keypoint heatmaps, which in turn provides certain guidance for the learning of offsets.
OA is hyper-parameter free, thus can be easily incorporated into any existing multi-task learning frameworks to boost the performance of main tasks.
We recommend using OA after $75\%$ training epochs when models start to converge.
\begingroup
\begin{figure}[pt]
\setlength{\tabcolsep}{2pt}
\centering
\begin{tabular}{cc}
\includegraphics[width=0.48\columnwidth]{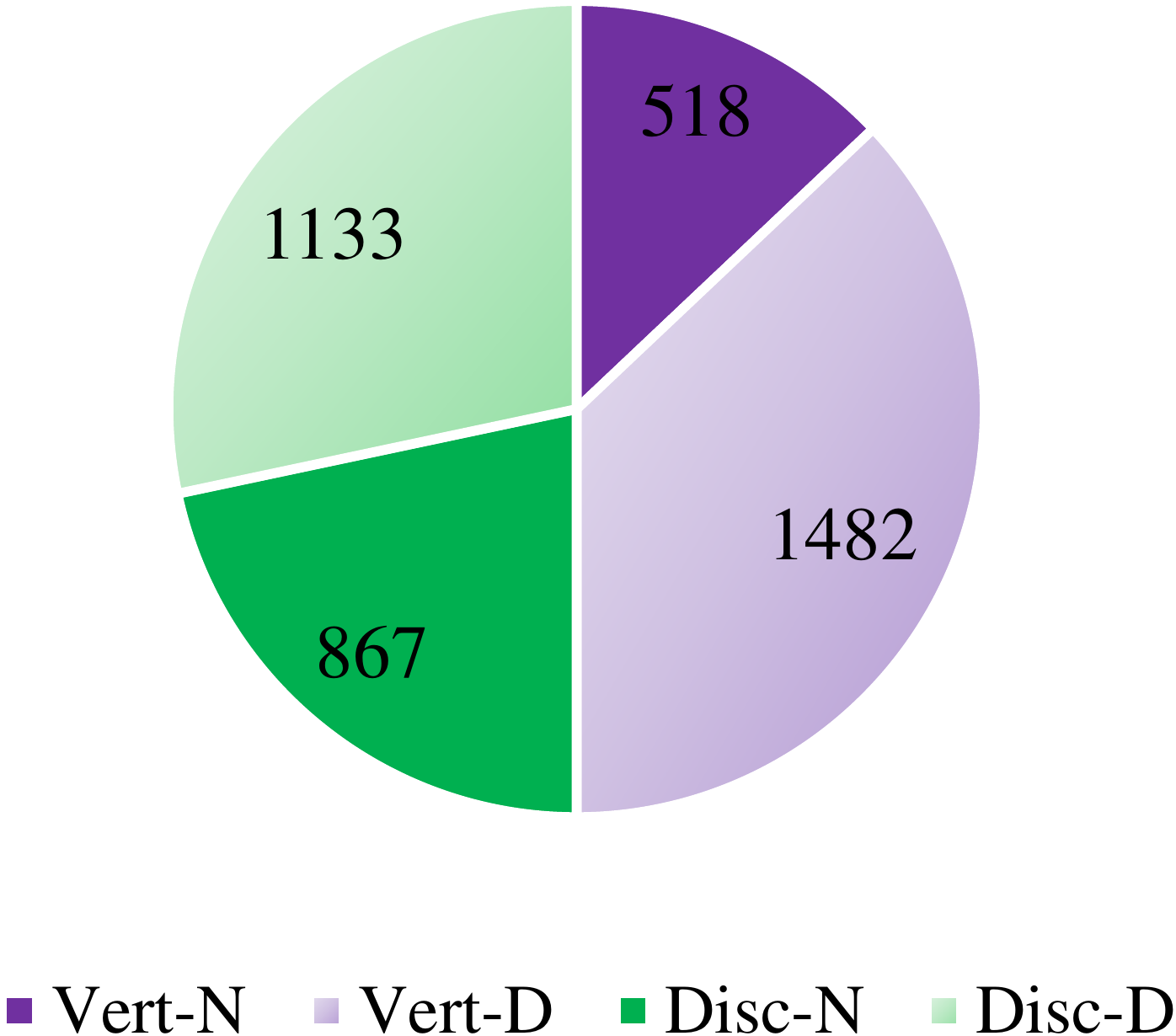}&
\includegraphics[width=0.48\columnwidth]{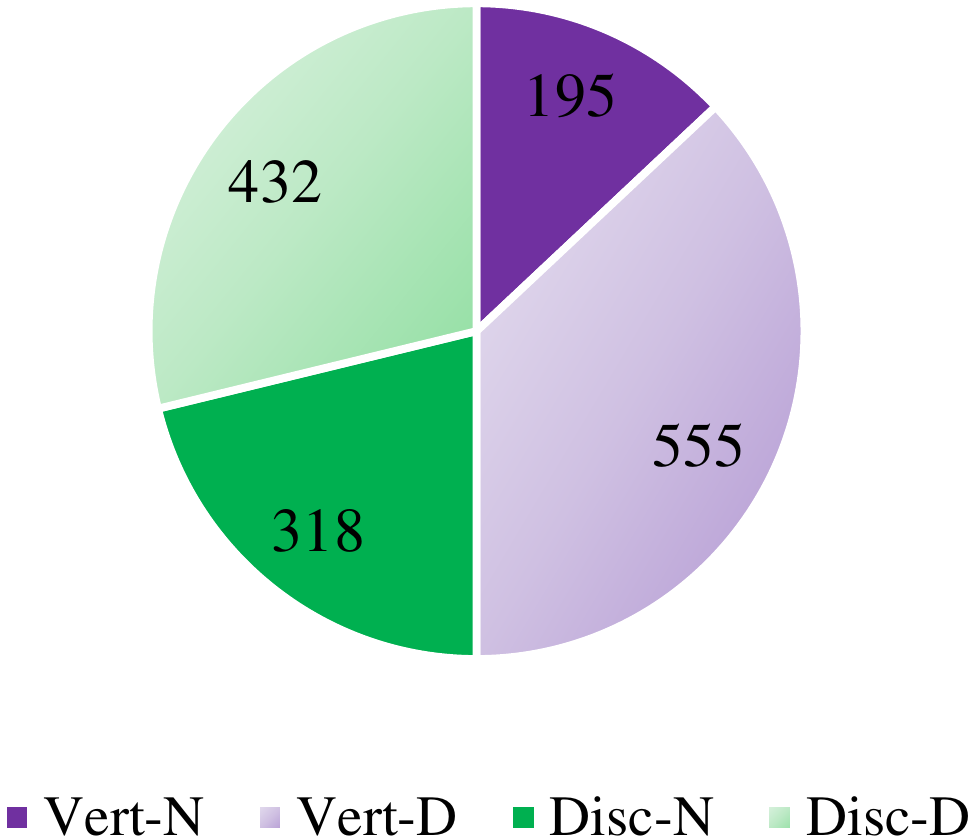}
\\
(a) Training set & (b) Test set
\end{tabular}
    \caption{Statistics of the SDID-TC dataset. Vert-N, Vert-D, Disc-N, and Disc-D denote the numbers of vertebrae being normal or degenerative, discs being normal or degenerative in the dataset, respectively.}
    \label{image:dataset}
\end{figure}
\endgroup

\section{Experiments}
\subsection{SDID-TC Dataset}
We evaluate our SpineOne framework on the Spinal Disease Intelligent Diagnosis Tianchi Competition (SDID-TC) dataset \cite{tianchi}, which inspired researchers to develop efficient and accurate AI frameworks for the purpose of solving medical problems.
We randomly split the 550 exams into 400 for training and 150 for test. The distribution of every class is shown in Fig. \ref{image:dataset}, with mild class imbalance in the dataset.
There are both T1 and T2 weighted sequences of MRIs in each exam. We only consider the sagittal T2 sequence as it highlights the cerebrospinal fluid (CSF) beside discs and vertebrae, which can help experts identify degenerative conditions more easily based on the brightly-shown deformation of discs and vertebrae, as well as their relative positions to spinal canals. To minimize the annotation work of experts, they were only required to annotate the centroids of discs and vertebrae on the middle slice of sagittal T2, as well as their corresponding degenerative conditions. This style of annotations saved a lot of time compared with annotations of bounding boxes or segmentation masks.

\subsection{Evaluation Metrics}
\label{sec:evaluation}
To evaluate our overall diagnosis performance (localization and classification), we first adopt the standard Percentage of Correct Keypoints (PCK) metric in keypoint localization tasks. PCK regards all generated keypoints that fall within a certain threshold distance to the ground truth as positive/detected ones. In our experiments, PCK under $6mm$ (PCK-$6$) is suggested by clinical experts as the threshold for positive discs and vertebrae. Among all positive centroids, correctly classified (i.e., correct pathological classification) ones are true positive (TP) while wrongly classified ones are false positive (FP). Those missed centroids are false negative (FN). After defining TP, FP, and FN, we further use them to compute the \textit{precision}, \textit{recall}, and \textit{F1 score} as the final evaluation for all frameworks comprehensively.

\subsection{Implementation Details}
A number of state-of-the-art methods are compared in our experiments. DeepLab series perform excellently in semantic segmentation, object detection, and panoptic segmentation \cite{chen2018encoder}. Here we use DeepLabv3+ with ResNet50/101 \cite{he2016deep} as the backbone, followed by the atrous spatial pyramid pooling (ASPP) consisting of a $1\times1$ convolution and three $3\times3$ convolutions with $rates=(6, 12, 18)$. DenseNet121/169 \cite{huang2017densely} is also compared as a backbone for global feature extraction, followed by feature pyramid networks (FPN) \cite{lin2017feature} as its neck. In addition, we also compare CenterNet (with ResNet101 as the backbone) \cite{zhou2019objects} and FCOS (with ResNet101 as the backbone) \cite{tian2019fcos} with their heads modified for our task. Different from above adapted one-stage methods, two-stage methods localize keypoints and then classify cropped regions around them subsequently. In order to demonstrate the superiority of SpineOne over two-stage methods, we utilize Hourglass \cite{newell2016stacked} and HRNet \cite{sun2019deep} for keypoint localization, followed by the ResNet18 as a classifier for those region proposals. Hourglass-$52$/$104$ denotes $1$ or $2$ hourglass modules. HRNet-W$32$/$48$ denotes different widths of the high-resolution subnetworks. We also tried deeper ResNets as classifiers but we did not obtain notable improvement.

We set dimensions as $H=W=128, \; C=64$ (Fig. \ref{image:attention_module}) for attention modules throughout all models. We set the batch size as $64$/$16$ for models without/with attention modules due to GPU memory constraints.
The initial learning rate is set to $0.04$/$0.01$ for different batch sizes and gradually decayed following $lr_i=(1-i/T)^{0.9}$, at the $i^{th}$ epoch with $T=300$ epochs in total. We use the Adam \cite{kingma2014adam} optimizer with default settings ($\beta_1=0.9$, $\beta_2=0.999$, $\epsilon=10^{-8}$). We select $\gamma=2$ in the focal loss \cite{lin2017focal}.
In the SDID-TC dataset, different exams may have different spacings/resolutions, which is the actual distance between two adjacent pixels in a slice. There are various resolutions ranging from $320\times 320$ to $640\times 640$, and various intervals ranging from $4.4mm$ to $4.6mm$. We resize all slices to $640\times 640$ pixels after spacing alignment to $0.4375mm \times 0.4375mm$ before feeding them into the network. Typical data augmentations including random horizontal flip, random zooming in/out ($[0.7, 1.3]$) and random crop (to the size of $512\times 512$ pixels) are also applied during training.
All models are trained using NVIDIA V100 GPUs without additional tricks.

\subsection{Results and Analysis}
Overall, SpineOne surpasses both one-stage and two-stage state-of-the-art methods by integrating all three proposed components into existing one-stage, anchor-free detection methods. Different components are able to boost each other to achieve the best performance across extensive one-stage models. SpineOne with DeepLabv3+(Res101) performs the best among all because ASPP is an excellent ingredient to enlarge the reception field, thus helping fuse multi-level features learned from the backbone. Moreover, SpineOne with CenterNet(Res101) also performs competitively well because CenterNet models an object as the centroid of its bounding box, which is a perfect design for our task when all anatomical structures are annotated as centroids. Visualization results are available in Fig. \ref{image:visualization_results}.
In addition, our framework can be easily generalized to other analogue tasks, especially medical diagnosis applications. For example, the simple yet effective OCPC heatmap can be applied to other localization and classification tasks where keypoints are relatively separated in space. Mapping keypoints into one channel can help capture the geometrical and classification correlations among different keypoints, which is particularly useful for tasks that require the exact location and classification of keypoints. Also, the two attention modules can be extended to tasks where different regions of inputs should be treated more adaptively. Furthermore, the gradient-based OA mechanism is also a potential technique to generally boost the learning of main objectives in multi-head models at the later training stage.

Two-stage methods need to train region proposal networks and classifiers in sequence. They are also more time-consuming during inference (e.g., our SpineOne is only $0.029$ seconds during inference on average, in contrast, two-stage methods cost more than $0.3$ seconds on average). However, it is surprising that two-stage methods do not outperform one-stage ones as empirically demonstrated in Table \ref{table:result0}. 
One important reason is that different from two-stage methods, one-stage methods are able to localize and classify all keypoints simultaneously, better leveraging their intrinsic correlations. For example, the degeneration of disc $D_2$ may also indicate the degeneration of adjacent discs $D_1$ and $D_3$, while both $D_4$ and $D_5$ are normal (Fig. \ref{spine}).
As a result, one-stage methods can take advantage of their geometrical and classification correlations as to predict their locations and classes more accurately than two-stage ones, which classify keypoints by cropping their surrounding regions separately. Our SpineOne can further improve the performance of one-stage methods integrating the OCPC keypoint heatmap, attention modules and the gradient-based OA mechanism.
\begin{table}[pt]
\caption{Performance comparison of our framework with the top three winning teams in SDID-TC. The score is the micro average precision of all categories with PCK under $6mm$.}
\begin{center}
\begin{tabular}{c|cccc}
\toprule
& \multicolumn{4}{c}{SDID-TC}\\
\midrule
Team (ranking) & $3^{rd}$ & $2^{nd}$ & $1^{st}$ & \textbf{SpineOne} \\
Score & 0.679 & 0.690 & 0.702 & \textbf{0.716} \\ 
\bottomrule
\end{tabular}
\label{table:competition}
\end{center}
\end{table}

Moreover, we also compare our results with the top three winning teams in SDID-TC \cite{tianchilist}. Two of them (the $1^{st}$ and $3^{rd}$ teams) converted the keypoint detection task into an object detection task by changing the ground truth annotations. Then, they exploited state-of-the-art detection methods to obtain good performance. The $2^{nd}$ team developed a one-stage detection framework for simultaneous localization and classification of degenerative discs and vertebrae, which is similar to our method. Table \ref{table:competition} reports the final score of SpineOne, which is higher than the best result in the competition. The score is the \textit{micro average precision}, defined as $score= \frac{TP}{TP+FP}$ of all categories with PCK under $6mm$. Again, we compute the \textit{macro mean} of \textit{precision}, \textit{recall}, and \textit{F1 score} towards overall evaluation of the final performance. These performance metrics allow the evaluation of our SpineOne and state-of-the-art methods in a more comprehensive manner.
\begin{table*}[pt]
\small
\caption{Ablation study of our framework on the SDID-TC dataset. The backbone architectures are DeepLabv3+(Res50/101). All evaluation results are the macro mean of normal and degenerative classes. The best results in every column are \textbf{boldfaced}.}
\begin{center}
\begin{tabular}{c|cccc|ccc|ccc}
\toprule
\textbf{Backbone} & \multicolumn{4}{c|}{\textbf{Component}} & \multicolumn{3}{c|}{\textbf{Discs}} & \multicolumn{3}{c}{\textbf{Vertebrae}} \\
& OCPC & PAM & CAM & OA & Recall & Precision & F1 & Recall & Precision & F1 \\
\midrule
\midrule
& &  &  &  & $0.813$ & $0.848$ & $0.830$ & $0.763$ &$0.784$ & $0.774$ \\
& $\checkmark$ &  &  &  & $0.827$ & $0.856$ & $0.841$ & $0.783$ &$0.797$ & $0.790$ \\
& $\checkmark$ & $\checkmark$ &  &  & $0.838$ & $0.865$ & $0.851$ & $0.790$ &$0.812$ & $0.801$ \\ 
DeepLabv3+ & $\checkmark$ &  & $\checkmark$ &  & $0.829$ & $0.859$ & $0.843$ & $0.785$ &$0.802$ & $0.793$ \\
(Res50) & $\checkmark$ & $\checkmark$ & $\checkmark$ &  & $0.841$ & $0.868$ & $0.854$ & $0.800$ &$0.823$ & $0.811$ \\
& & $\checkmark$ & $\checkmark$ & & $0.823$ & $0.851$ & $0.836$ & $0.776$ &$0.792$ & $0.784$ \\
& $\checkmark$ &  &  & $\checkmark$ & $0.832$ & $0.861$ & $0.846$ & $0.785$ &$0.806$ & $0.795$ \\
& $\checkmark$ & $\checkmark$ & $\checkmark$ & $\checkmark$ & $\textbf{0.847}$ & $\textbf{0.873}$ & $\textbf{0.860}$ & $\textbf{0.811}$ & $\textbf{0.834}$ & $\textbf{0.823}$ \\ 
\midrule
& &  &  &  & $0.819$ & $0.855$ & $0.836$ & $0.774$ & $0.795$ & $0.785$ \\
& $\checkmark$ &  &  &  & $0.836$ & $0.870$ & $0.852$ & $0.796$ &$0.816$ & $0.806$ \\
& $\checkmark$ & $\checkmark$ &  &  & $0.850$ & $0.882$ & $0.865$ & $0.816$ &$0.838$ & $0.827$ \\ 
DeepLabv3+ & $\checkmark$ &  & $\checkmark$ &  & $0.847$ & $0.878$ & $0.861$ & $0.810$ &$0.832$ & $0.821$ \\
(Res101) & $\checkmark$ & $\checkmark$ & $\checkmark$ &  & $0.853$ & $0.885$ & $0.868$ & $0.826$ &$0.848$ & $0.837$ \\
& & $\checkmark$ & $\checkmark$ & & $0.830$ & $0.864$ & $0.846$ & $0.790$ &$0.810$ & $0.800$ \\
& $\checkmark$ &  &  & $\checkmark$ & $0.844$ & $0.875$ & $0.859$ & $0.805$ &$0.827$ & $0.816$ \\
& $\checkmark$ & $\checkmark$ & $\checkmark$ & $\checkmark$ & $\textbf{0.859}$ & $\textbf{0.891}$ & $\textbf{0.874}$ & $\textbf{0.840}$ & $\textbf{0.859}$ & $\textbf{0.849}$ \\  
\bottomrule
\end{tabular}
\label{table:result1}
\end{center}
\end{table*}

\begin{table*}[pt]
\small
\caption{Performance comparison between w/ (even rows) and w/o (odd rows) the offset map as outputs.}
\begin{center}
\begin{tabular}{c|cccc|ccc|ccc}
\toprule
\textbf{Backbone} & \multicolumn{4}{c|}{\textbf{Component}} & \multicolumn{3}{c|}{\textbf{Discs}} & \multicolumn{3}{c}{\textbf{Vertebrae}} \\
& Offset & OCPC & PAM & CAM & Recall & Precision & F1 & Recall & Precision & F1 \\
\midrule
\midrule
& & $\checkmark$ &  &  & $0.824$ & $0.853$ & $0.838$ & $0.774$ &$0.796$ & $0.785$ \\
DeepLabv3+ & $\boldcheckmark$ & $\checkmark$ &  &  & $0.827$ & $0.856$ & $0.841$ & $0.783$ &$0.797$ & $0.790$ \\
\cline{2-11}
(Res50) & & $\checkmark$ & $\checkmark$ & $\checkmark$ & $0.837$ & $0.862$ & $0.849$ & $0.792$ & $0.817$ & $0.804$ \\
& $\boldcheckmark$ & $\checkmark$ & $\checkmark$ & $\checkmark$ & $0.841$ & $0.868$ & $0.854$ & $0.800$ &$0.823$ & $0.811$ \\
\midrule
& & $\checkmark$ &  &  & $0.833$ & $0.861$ & $0.846$ & $0.791$ & $0.816$ & $0.803$ \\
DeepLabv3+ & $\boldcheckmark$ & $\checkmark$ &  &  & $0.836$ & $0.870$ & $0.852$ & $0.796$ &$0.816$ & $0.806$ \\
\cline{2-11}
(Res101) & & $\checkmark$ & $\checkmark$ & $\checkmark$ & $0.849$ & $0.874$ & $0.861$ & $0.820$ &$0.843$ & $0.831$ \\
& $\boldcheckmark$ & $\checkmark$ & $\checkmark$ & $\checkmark$ & $0.853$ & $0.885$ & $0.868$ & $0.826$ &$0.848$ & $0.837$ \\
\bottomrule
\end{tabular}
\label{table:offset}
\end{center}
\end{table*}

\subsection{Ablation Studies}
Table \ref{table:result1} reports the experimental results of our framework under various settings: 1) different backbones; and 2) with different components (i.e., the OCPC keypoint heatmap, PAM and CAM attention modules, and the gradient-based OA mechanism). 
\begingroup
\begin{figure}[pt]
\setlength{\tabcolsep}{2pt}
\centering
\begin{tabular}{cc}
\includegraphics[width=0.48\columnwidth]{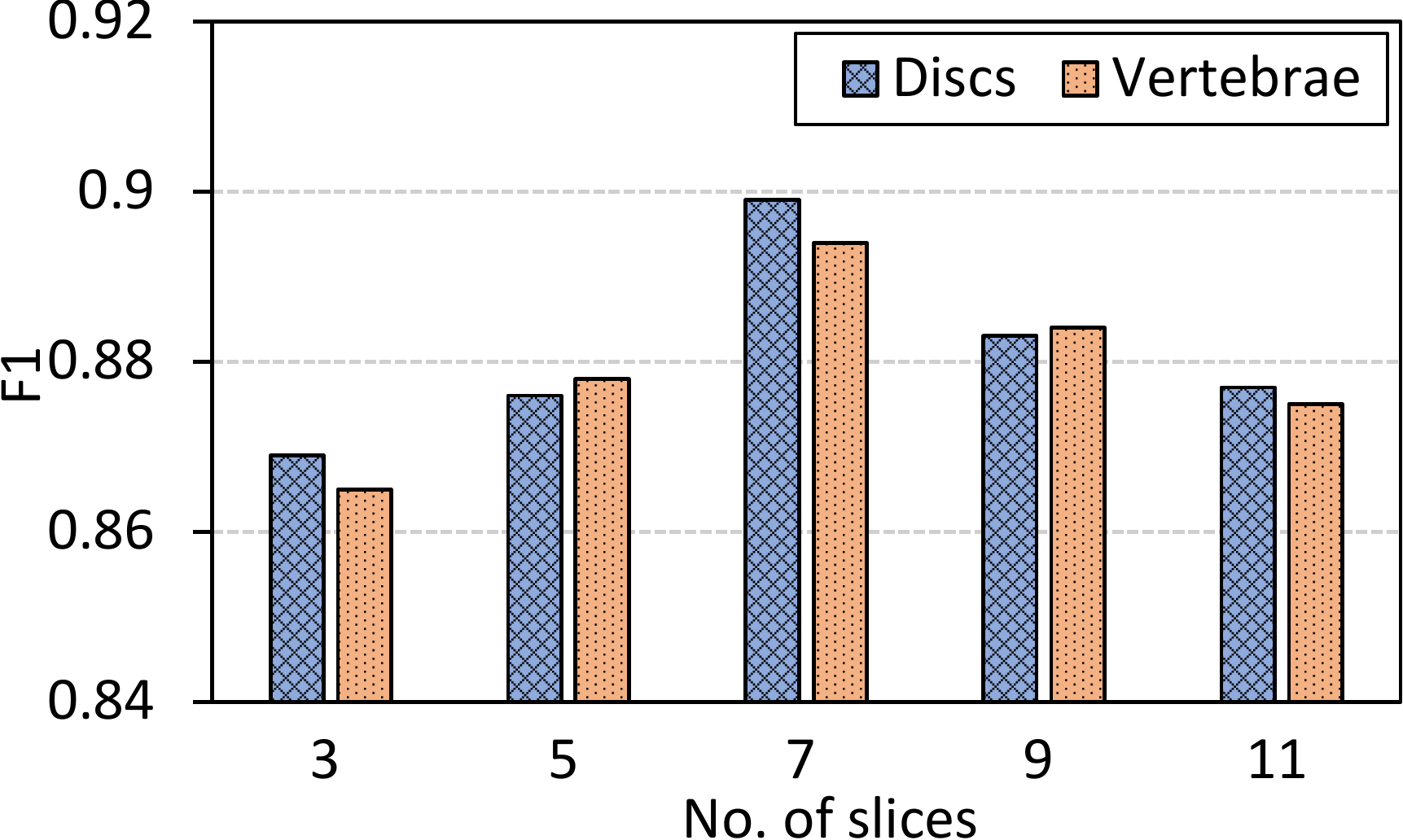}&
\includegraphics[width=0.48\columnwidth]{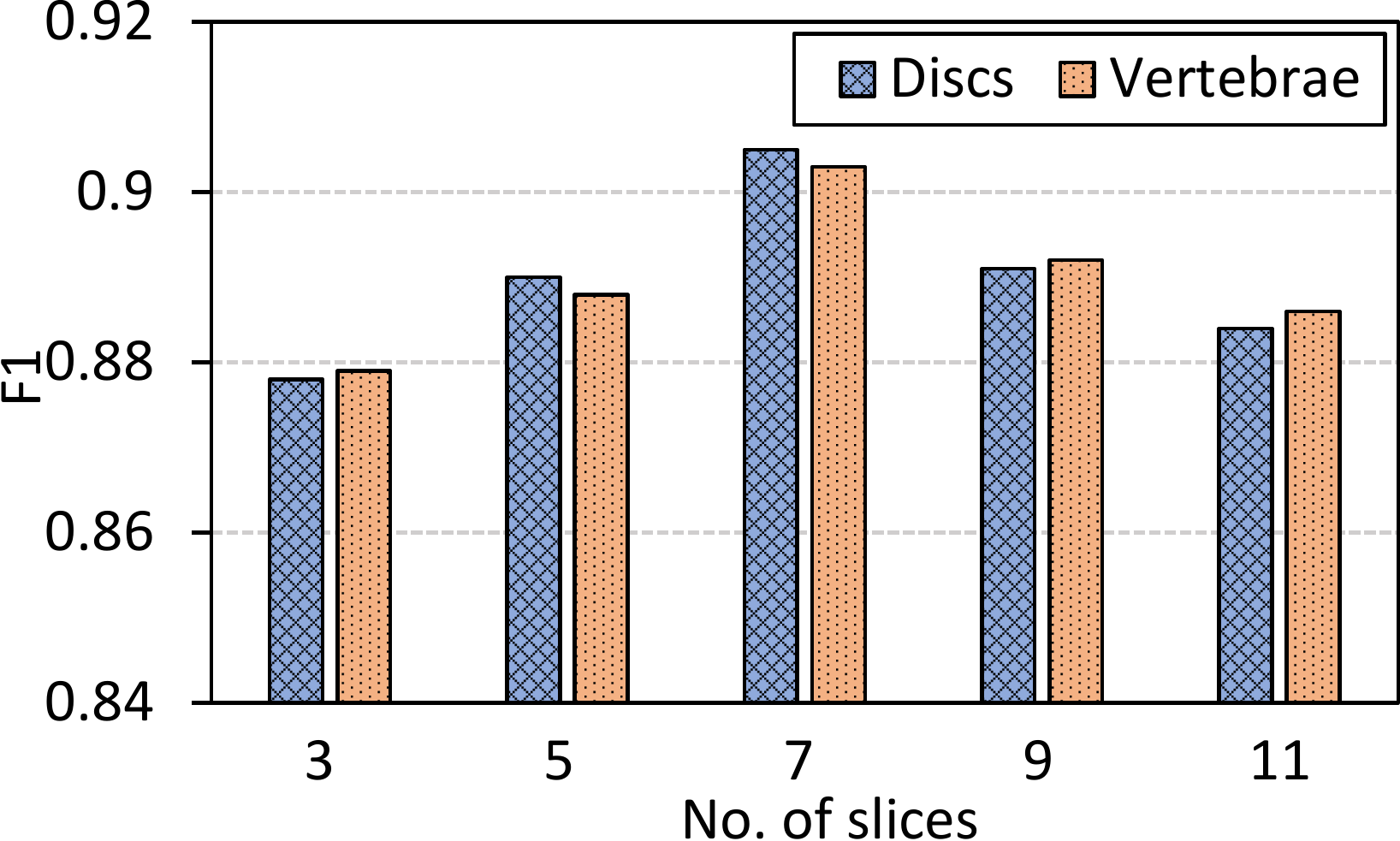}
\\
(a) DeepLabv3+(Res50) & (b) DeepLabv3+(Res101)
\end{tabular}
    \caption{Comparison of using different numbers of MRI slices as inputs.}
    \label{image:slices}
\end{figure}
\endgroup

As can be observed, each of our proposed component can improve stage-of-the-art one-stage methods, and our integrated SpineOne framework outperforms them by $3\%-6.4\%$ (absolute) in terms of macro mean F1, to which precision contributes more than recall.
Note that the conventional heatmap will have $7$ channels ($5$ centroids and $2$ classes) for discs and vertebrae without OCPC, respectively. Our simple OCPC design with only $2$ channels ($2$ classes) can help improve the performance by approximately $2\%$.
PAM performs better than CAM dominantly, which indicates that the inner-image spatial interaction is more important than the inter-channel correlation in our task.
The gradient-based OA mechanism can also help improve the final performance of our framework with or without attention modules.
Note that the framework with attention modules does not perform well without the OCPC heatmap, which proves the necessity of the OCPC design when diagnosing degenerative discs and vertebrae.

\subsection{More Explorations}
\subsubsection{Slices selection}
There are several sequences of MRIs in each exam, e.g., sagittal T1 weighted, sagittal T2 weighted, and axial T2 weighted. The cerebrospinal fluid (CSF) is highlighted (brightly shown) in T2 sequences whereas dark in T1 sequences. Clinical experts mainly focus on the slices in the middle of sagittal T2 to diagnose degenerative discs and vertebrae \cite{jamaludin2016spinenet}. 
As we explained earlier, this is also the reason why we take the slice in the middle of sagittal T2 and its nearby ones as inputs. However, there are usually $11$ or more slices in sagittal T2. Therefore, we conduct a comparison study to test the performance on different number of input slices (i.e., slices in number of $[3,5,7,9,11]$). 
This is done with OCPC, but without attention modules or OA for the convenience of comparison and avoiding the effect of other factors.
As shown in Fig. \ref{image:slices}, the performance is consistent across the two backbones and the best results are obtained when $7$ slices in the middle of sagittal T2 are considered. This indicates that slices in the middle of sagittal T2 can indeed help models identify degenerative discs and vertebrae, while those far away from the middle are less useful.

\subsubsection{With or without the offset}
Since the offset learning is less important than the keypoint localization and classification, here we further test the performance of our framework with or without the two (i.e., discs and vertebrae) offset outputs, with results reported in Table \ref{table:offset}. 
The short-range offset map uses pixels around the keypoint as the supplementary information to locate it. Note that our OA mechanism has to be removed in this case as it is designed to better associate the offset with the heatmap.
It can be observed that models with the offset map (even rows) can generally perform better than those without the offset map (odd rows), to limited extent (less than $1\%$ improvement on the F1 score in Table \ref{table:offset}). 
As empirically proved by the results in Table \ref{table:result1}, when the offset is considered, our OA mechanism can further improve the performance by approximately $1\%$ on average.

\subsubsection{Detection rate}
\begin{figure}[t]
\centering
\includegraphics[width=0.85\columnwidth]{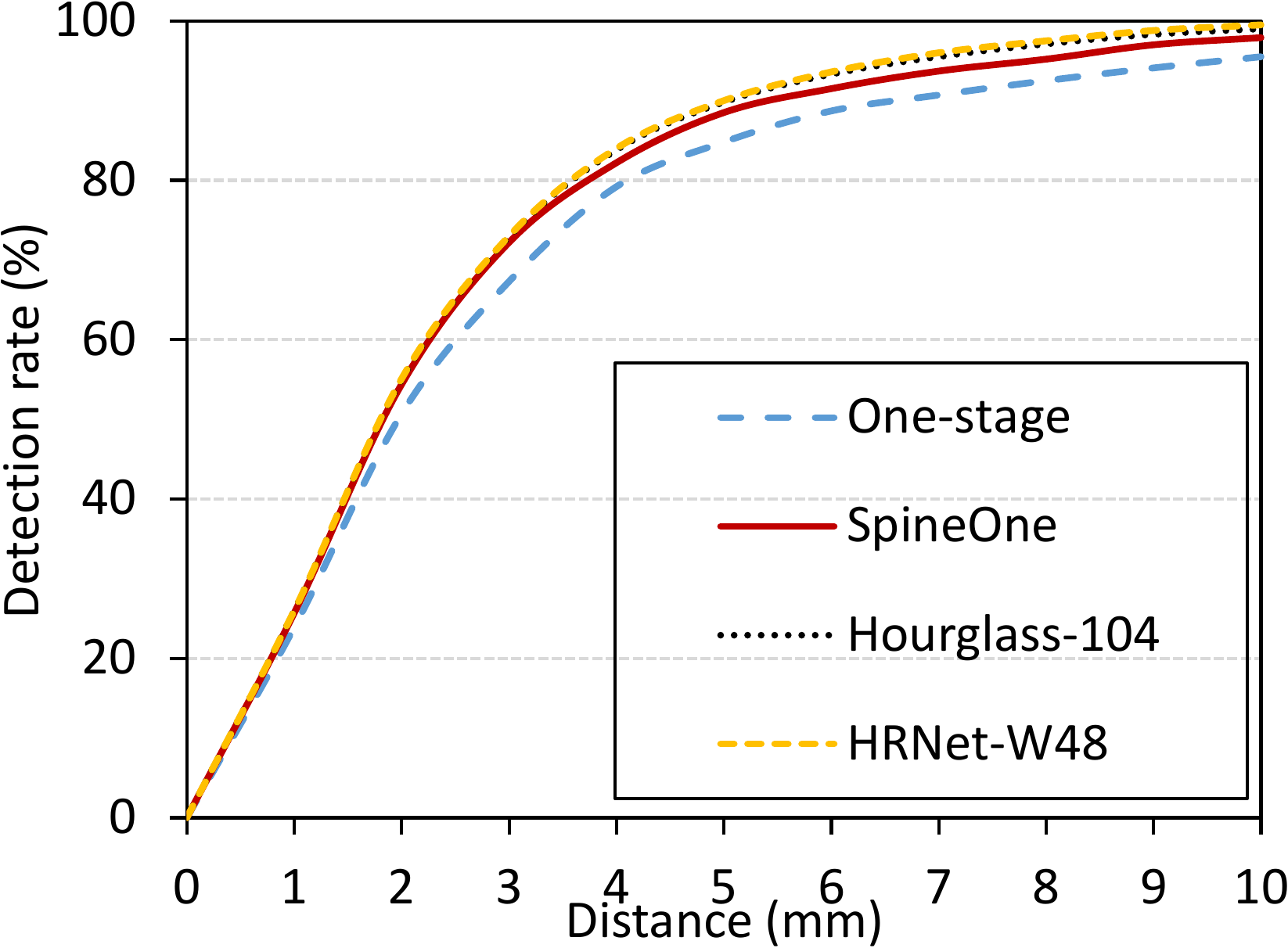}
\caption{PCK comparison results. We select DeepLabv3+(Res101) as the backbone for both the conventional one-stage method and SpineOne. We select Hourglass-104 and HRNet-W48 as examples of two-stage methods.}
\label{image:pck}
\end{figure}
Although the localization accuracy is not our final goal, it is still worth inspecting it to better understand our framework. Here, we conduct the last experiment to evaluate models by only the detection/identification rate of all centroids, which is calculated by the standard Percentage of Correct Keypoints (PCK) metric in keypoint detection tasks. PCK can be formulated in our task as follows:
\begin{equation}
\centering
\small
PCK = \frac{TP+FP}{TP+FP+FN}.
\label{fun:evaluation}
\end{equation}

PCK under $6mm$ (PCK-$6$) was introduced in Section \ref{sec:evaluation} suggested by experts. Here, we calculate the detection rate by PCK in an extensive range from $1mm$ to $10mm$ (Fig. \ref{image:pck}).
We compare four models, i.e., the one-stage method with DeepLabv3+(Res101), SpineOne with DeepLabv3+(Res101), Hourglass-104, and HRNet-W48, which are $2^{nd}$, $4^{th}$, $9^{th}$, and $15^{th}$ models in Table \ref{table:result0}.
Although there are some variations, our proposed framework demonstrates an approximately $2-3\%$ improvement over the one-stage method across the extensive range. For example, PCK-$6$ of the one-stage method is improved from $88.7\%$ to $91.5\%$, while PCK-$10$ is improved from $95.5\%$ to $97.9\%$ by our SpineOne. We also observe that the improvement of the detection rate evaluated by PCK is less significant than those results reported in Table \ref{table:result0}. The reason is that the proposed OA mechanism encourages our framework to focus on the more important keypoint learning task, rather than the offset learning task. Indeed, the pathological classification in the keypoint learning task plays a more important role in spinal diagnosis than only finding the optimal location of the discs or vertebrae by offset learning. We could actually tolerate some small errors in centroids localization, which is why PCK-$6$ was set by experts. Interestingly, two-stage methods outperform SpineOne in keypoint localization by about $1\%$ across the extensive range. However, it drops much in classification as it cropped the entire input into small region proposals based on centroids' locations, which fails to capture the classification correlation among discs and vertebrae.
\begin{table}[tp]
\caption{Performance comparison of our framework with existing baselines in the MICCAI 2014 Computational Challenge on Vertebrae Localization and Identification.}
\scriptsize
\centering
\begin{tabular}{c|ccccc}
\toprule
Independent & Chen & Yang & Liao & Chen & \multirow{2}{*}{Ours} \\
test & et al. \cite{chen2015automatic} & et al. \cite{yang2017automatic} & et al. \cite{liao2018joint} & et al. \cite{chen2019vertebrae} & \\
\midrule
Detection rate & $84.16\%$ & $85.00\%$ & $88.30\%$ & $94.67\%$ & $94.55\%$ \\ 
\bottomrule
\end{tabular}
\label{table:MICCAI}
\end{table}

In addition, we also compare our method with existing baselines on a public dataset from the MICCAI 2014 Computational Challenge on Vertebrae Localization and Identification (Table \ref{table:MICCAI}), which is not exactly the same task but close to ours. The dataset contains 242 spine-focused CT training scans of various types of high-grade pathologies and metal implants, together with 60 other scans for hold-out evaluation. The differences between this task and our task are two-fold: 1) the input images are 3D CT volumes in this task while the inputs are 2D MRI slices in our task; 2) there is no degeneration classification in this task, so we only use the detection rate as the evaluation metric. We modify the structure of our framework so that it can take as input the 3D CT volumes ($512\times 512\times 128$). We empirically show that our method achieves the state-of-the-art performance on this public dataset, where the baseline results are directly copied from their original papers.
\begin{table*}[tp]
\caption{Structure of our framework. Conv2d() denotes Conv2d(in\_channel, out\_channel, kernel, stride). BN denotes batch normalization.}
\begin{center}
\begin{tabular}{c|c|c|c|c}
\toprule
\textbf{Modules} & \multicolumn{4}{c}{\textbf{Parameters}} \\
\midrule
\midrule
Backbone & \multicolumn{4}{c}{e.g., DeepLabv3+} \\
\midrule
Two substreams & \multicolumn{2}{c|}{Discs} & \multicolumn{2}{c}{Vertebrae} \\
\midrule
Attention & \multicolumn{2}{c|}{PAM+CAM} & \multicolumn{2}{c}{PAM+CAM} \\
\midrule
\multirow{4}{*}{CNN} & Conv2d(64, 32, 3, 1) & Conv2d(64, 32, 3, 1) & Conv2d(64, 32, 3, 1) & Conv2d(64, 32, 3, 1) \\
& BN & BN & BN & BN \\
& Conv2d(32, 32, 3, 1) & Conv2d(32, 32, 3, 1) & Conv2d(32, 32, 3, 1) & Conv2d(32, 32, 3, 1) \\
& BN & BN & BN & BN \\
\midrule
Generation & Conv2d(32, 2, 1, 1) & Conv2d(32, 4, 1, 1) & Conv2d(32, 2, 1, 1) & Conv2d(32, 4, 1, 1) \\
\midrule
Output & Heatmap $\bm{\hat{h}}_{disc}$ & Offset $\bm{\hat{o}}_{disc}$ & Heatmap $\bm{\hat{h}}_{vert}$ & Offset $\bm{\hat{o}}_{vert}$ \\
\bottomrule
\end{tabular}
\label{table:framework}
\end{center}
\end{table*}

\begingroup
\begin{figure*}[tp]
\setlength{\tabcolsep}{2pt}
\centering
\begin{tabular}{cc}
\includegraphics[width=\columnwidth]{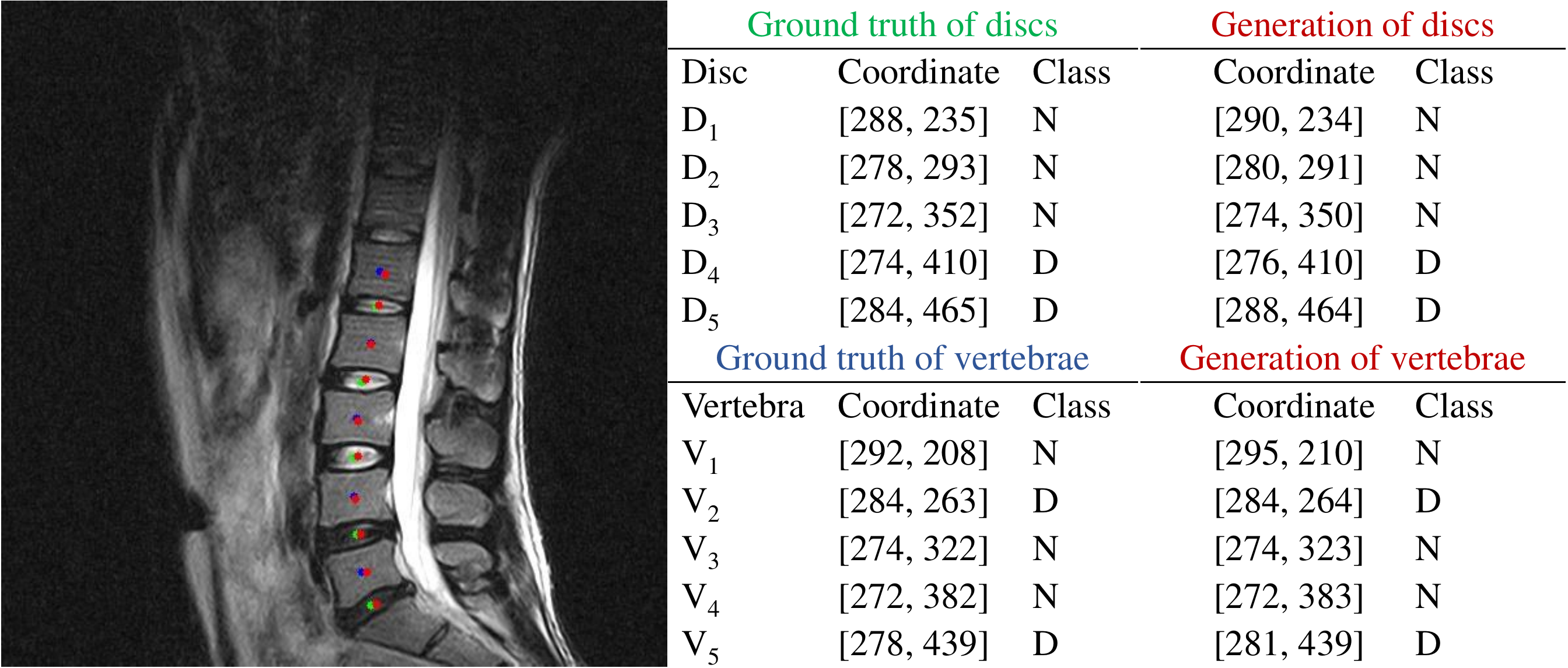}&
\includegraphics[width=\columnwidth]{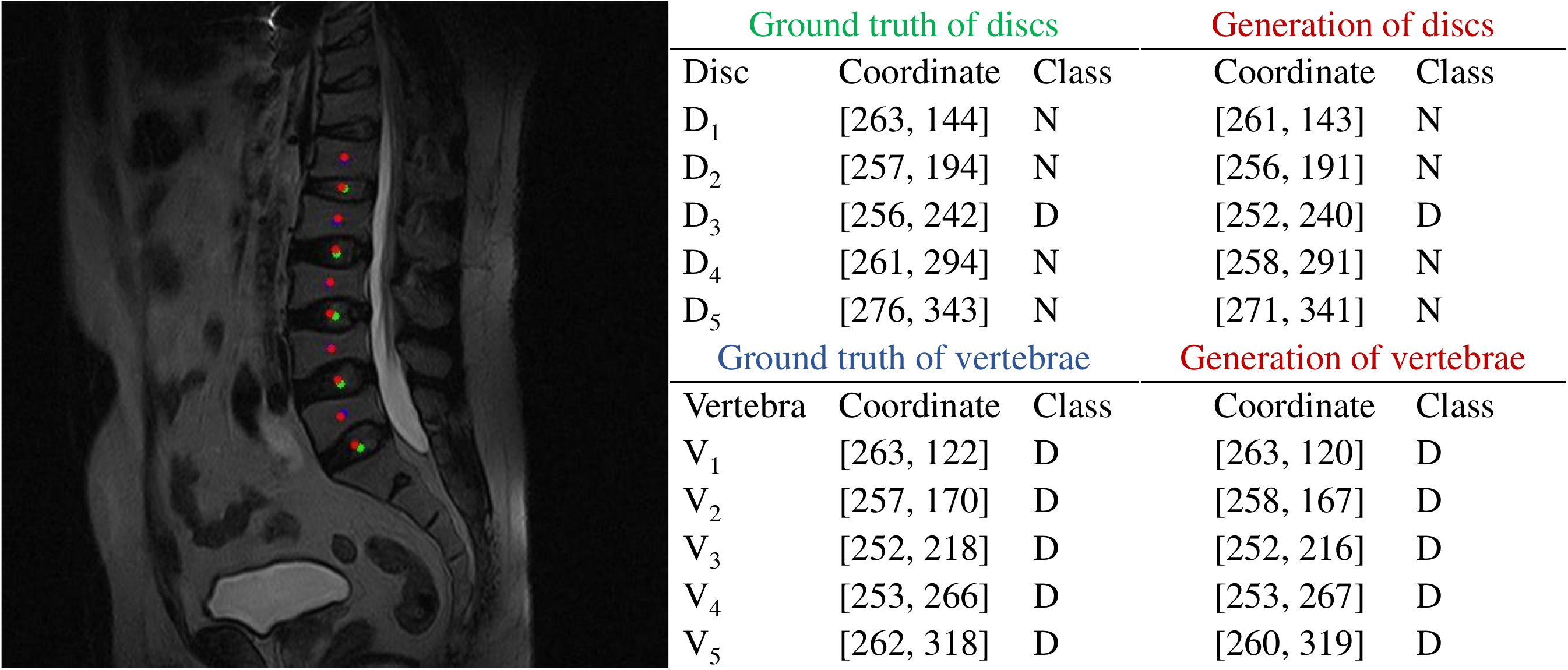} \\
\includegraphics[width=\columnwidth]{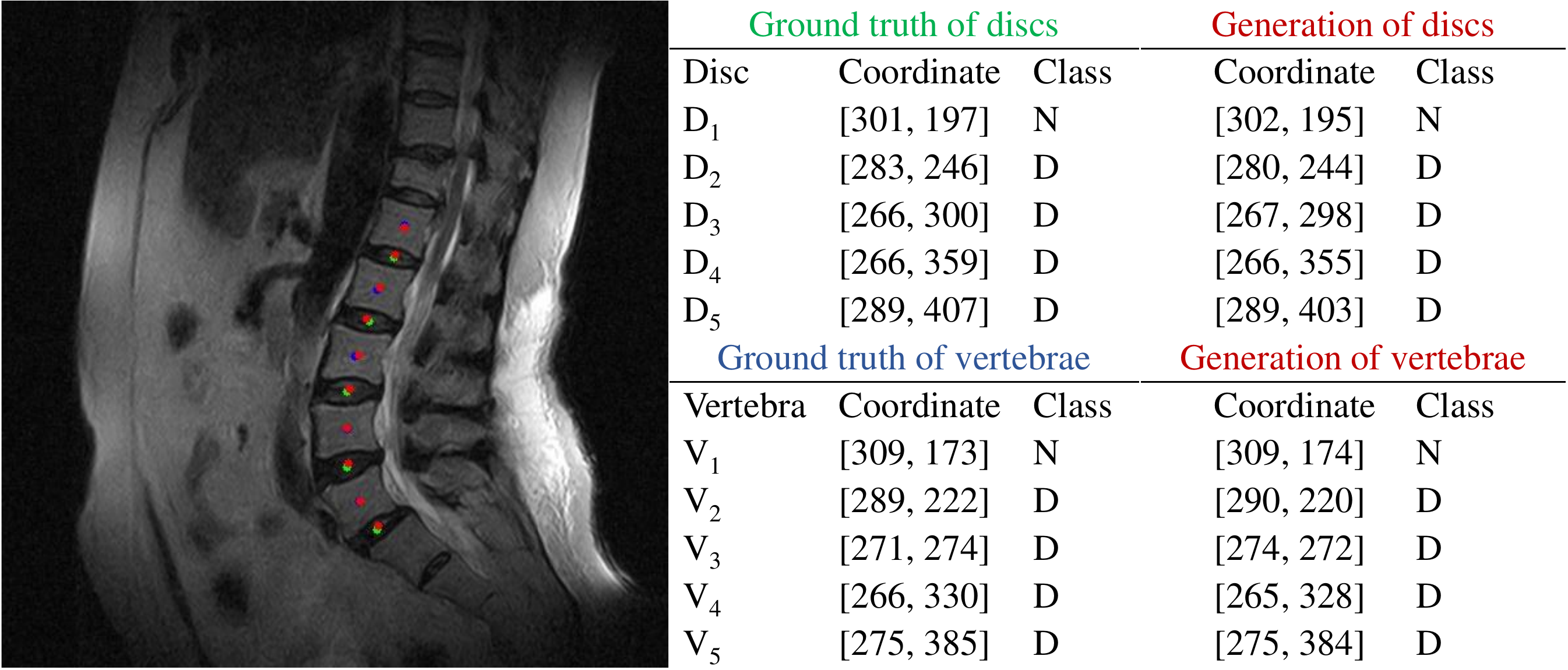}&
\includegraphics[width=\columnwidth]{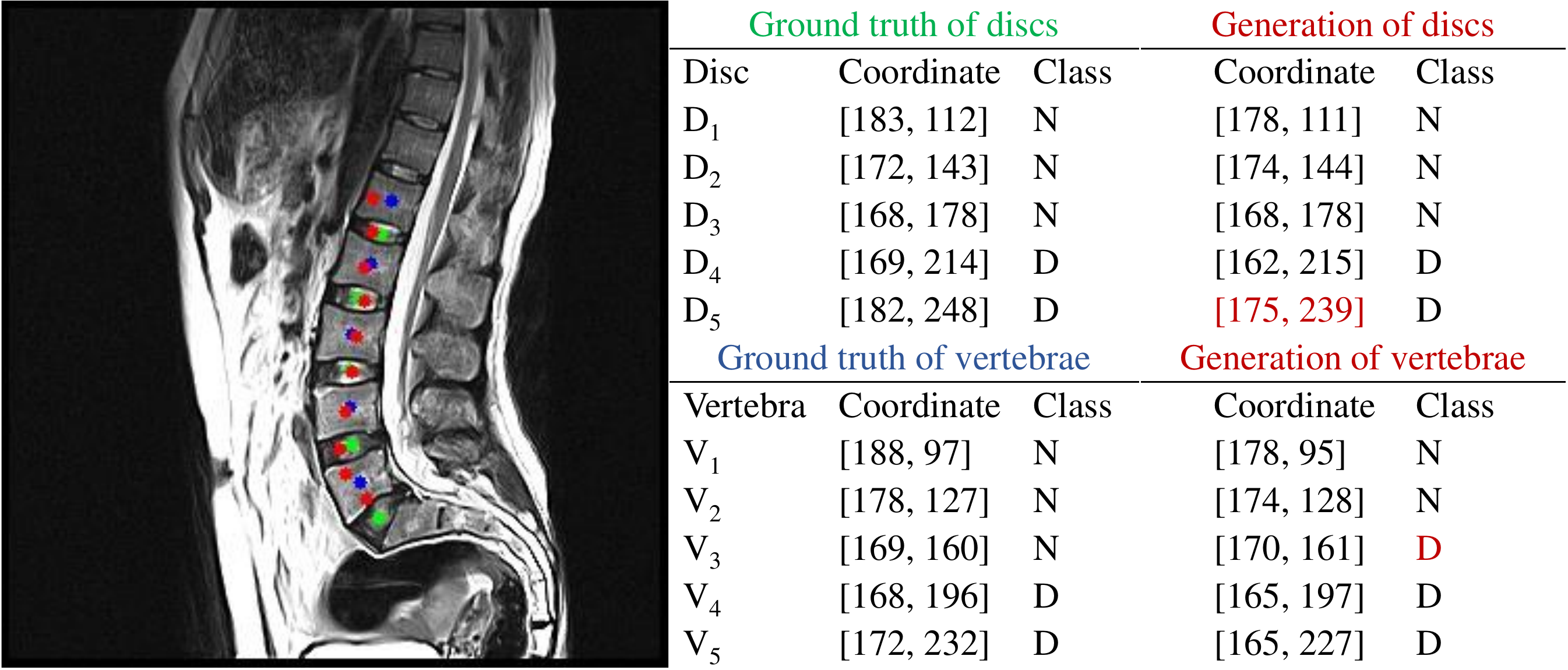}
\end{tabular}
    \caption{Random inference examples of generated results by our framework, compared with the ground truth. In each case, the left image is the middle slice with generated (red) and ground truth centroids of discs (green) and vertebrae (blue). The middle/right table shows ground truth/generated pixel locations and degenerative classes. The first three examples are successful cases while the last (bottom right) one is a failure case. Our approach demonstrates good performance in most cases when input slices of the sagittal T2 sequence are with major spacings among the dataset. However, when inputs are with few spacings among the dataset (e.g., slices in the last case are with the extremely large spacing, namely the low resolution), our framework may detect centroids not accurately or misclassify them into wrong classes. The black results in the four tables on the right are good results while the marked results in red are bad ones.}
    \label{image:visualization_results}
\end{figure*}
\endgroup

\section{Conclusion}
We propose a one-stage detector to simultaneously localize and classify degenerative discs and vertebrae on the lumbar spine. It addresses two essential tasks of the diagnosis process: 1) anatomical localization; and 2) pathological classification. SpineOne is built upon CNNs with three novel and generic techniques. First, a new design of the OCPC heatmap facilitates above two tasks at the same time. OCPC can help capture both geometrical and classification correlations among keypoints, and can be easily applied to other medical domains where keypoints are spatially separated. Second, dual self-attention modules better differentiate the learning stream for discs and vertebrae, which can be readily plugged into other frameworks to improve discriminative representation learning. Third, a novel gradient-guided OA mechanism further associates different learning objectives. It explicitly connects heatmap learning with offset learning, thus making the learning of the main objective more effective. We recommend using OA after $75\%$ training epochs when models start to converge. Experimental results show that we can improve state-of-the-art methods using each individual technique and the integrated framework wins by a large margin on the SDID-TC dataset. For future work, we plan to 1) augment the volume of the SDID-TC dataset; 2) research on the degenerative grading for discs and vertebrae; and 3) verify our three proposed techniques in other medical scenarios.

\section*{Acknowledgment}
We would like to thank clinical experts who have helped build the Spinal Disease Intelligent Diagnosis Tianchi Competition (SDID-TC) dataset. We are grateful for anonymous patients who have provided their MRI exams for our research. We are also thankful for the technical support from our colleagues at the DAMO Academy, Alibaba Group.

\bibliographystyle{IEEEtran}
\bibliography{IEEEabrv,main}

\end{document}